\documentclass[11pt,a4paper]{article}
\usepackage{hyperref}
\usepackage{acl2020}
\usepackage{times}
\usepackage{mathabx}
\usepackage{latexsym}
\usepackage{todonotes}
\usepackage{booktabs}
\usepackage{graphicx}
\usepackage{enumitem}
\usepackage{comment}

\usepackage{tikz}
\usepackage{tikz-dependency}
\usetikzlibrary{%
  shapes,%
  arrows,%
  positioning,%
  calc,%
  automata%
}

\usepackage{microtype}

\usepackage{algorithm}
\usepackage{algpseudocode}
\usepackage{dblfloatfix}
\algtext*{EndIf}
\algtext*{EndFor}
\aclfinalcopy 

\title{Do Neural Language Models Show Preferences for Syntactic Formalisms?}

\author{Artur Kulmizev \\
  Uppsala University\\
  \texttt{artur.kulmizev@lingfil.uu.se} \\\And
   Vinit Ravishankar\\
  University of Oslo \\
  \texttt{vinitr@ifi.uio.no} \\\AND
    Mostafa Abdou\\
  University of Copenhagen \\
  \texttt{abdou@di.ku.dk} \\\And
  Joakim Nivre\\
  Uppsala University \\
  \texttt{joakim.nivre@lingfil.uu.se} \\}

\date{}

\begin{document}
\maketitle
\begin{abstract}

Recent work on the interpretability of deep neural language models has concluded that many properties of natural language syntax are
encoded in their representational spaces. However, such studies
often suffer from limited scope
by focusing on a single language and a single linguistic formalism. In this study, we aim to investigate the extent to which the semblance of syntactic structure captured by language models adheres to a surface-syntactic or deep syntactic style of analysis, and whether the patterns are consistent across different languages. 
We apply a probe for extracting directed dependency trees 
to BERT and ELMo models trained on 13 different languages, probing for two different syntactic annotation styles: Universal Dependencies (UD), prioritizing deep syntactic relations, and Surface-Syntactic Universal Dependencies (SUD), focusing on surface structure. We find that both models exhibit a preference for UD over SUD --- with interesting variations across languages and layers --- and that the strength of this preference is correlated with differences in tree shape. 

\end{abstract}

\section{Introduction}

Recent work on interpretability in NLP has led to the consensus that deep neural language models trained on large, unannotated datasets manage to encode various aspects of syntax as a byproduct of the training objective. Probing approaches applied to models like ELMo \citep{peters-etal-2018-deep} and BERT \citep{devlin-etal-2019-bert} have demonstrated that one can decode various linguistic properties such as part-of-speech categories, dependency relations, and named-entity types directly from the internal hidden states of a pretrained model \citep{tenney_what_2019,tenney_what_2019,peters-etal-2018-dissecting}. Another line of work has tried to tie cognitive measurements or theories of human linguistic processing to the machinations of language models, often establishing strong parallels between the two \citep{prasad-etal-2019-using,abnar-etal-2019-blackbox,gauthier-levy-2019-linking}.

As is the case for NLP in general, English has served as the de facto testing ground for much of this work, with other languages often appearing as an afterthought. However, despite its ubiquity in the NLP literature, English is generally considered to be atypical across many typological dimensions. 
Furthermore, the tendency of interpreting NLP models with respect to existing, canonical datasets often comes with the danger of conflating the theory-driven annotation therein with scientific fact. One can observe this to an extent with the Universal Dependencies (UD) project \cite{nivre-etal-2016-universal}, which aims to collect syntactic annotation for a large number of languages. Many interpretability studies have taken UD as a basis for training and evaluating probes, 
but often fail to mention that UD, like all annotation schemes, is built upon specific theoretical assumptions, which may not be universally accepted.
Our research questions start from these concerns. When probing language models for syntactic dependency structure, is UD --- with its emphasis on syntactic relations between content words --- really the best fit? Or is the representational structure of such models better explained by a scheme that is more oriented towards surface structure, such as the recently proposed Surface-Syntactic Universal Dependencies (SUD) \cite{gerdes18udw}? And are these patterns consistent across typologically different languages?
To explore these questions, we fit the structural probe of \citet{hewitt2019structural} on pretrained BERT and ELMo representations, supervised by UD/SUD treebanks for 13 languages, and extract directed dependency trees. We then conduct an extensive error analysis of the resulting probed parses, in an attempt to qualify our findings. Our main contributions are the following:

\begin{enumerate}[topsep=2pt,noitemsep]
    \item 
    A simple algorithm for deriving directed trees from the disjoint distance and depth probes introduced by \citet{hewitt2019structural}.
    \item 
    A multilingual analysis of the probe's performance across 13 different treebanks.
    \item 
    An analysis showing that the 
    syntactic information encoded by BERT and ELMo 
    fit UD better than SUD for most languages.
\end{enumerate}

\begin{figure*}
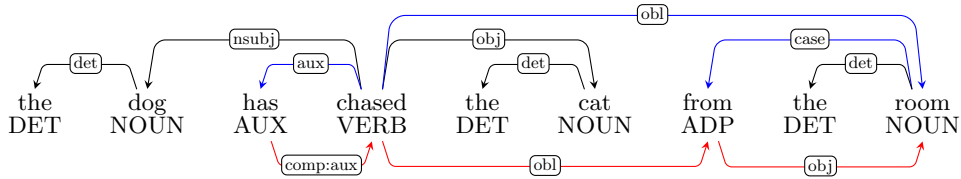

\small
\centering
\begin{dependency}
    \tikzset{every node}=[\fontfamily{palatino}]
\begin{deptext}[column sep=0.5cm]
the \& dog \& has \& chased \& the \& cat \& from \& the \& room \\
\textcolor{black}{DET} \& \textcolor{black}{NOUN} \& \textcolor{black}{AUX} \& \textcolor{black}{VERB} \& \textcolor{black}{DET} \& \textcolor{black}{NOUN} \& \textcolor{black}{ADP} \& \textcolor{black}{DET} \& \textcolor{black}{NOUN} \\
\end{deptext}
\depedge[edge unit distance=1em]{2}{1}{det}
\depedge[edge unit distance=1em]{4}{2}{nsubj}
\depedge[edge unit distance=1em, edge style=blue]{4}{3}{aux}
\depedge[edge unit distance=1em, edge style=red, edge below]{3}{4}{comp:aux}
\depedge[edge unit distance=1em]{6}{5}{det}
\depedge[edge unit distance=1em]{4}{6}{obj}
\depedge[edge unit distance=1em, edge style=blue]{9}{7}{case}
\depedge[edge unit distance=1em]{9}{8}{det}
\depedge[edge unit distance=0.6em, edge style=blue]{4}{9}{obl}
\depedge[edge style=red, edge below, edge unit distance=0.34em]{4}{7}{obl}
\depedge[edge style=red, edge below, edge unit distance=0.52em]{7}{9}{obj}
\end{dependency}
\caption{Simplified \textcolor{blue}{UD} and \textcolor{red}{SUD} annotation for an English sentence.}
\label{fig:en-fi}
\end{figure*}

\section{Related Work}
    
There has been a considerable amount of recent work attempting to 
understand what aspects of natural language pre-trained encoders 
learn. The classic formulation of these probing experiments is in the form of diagnostic classification \citep{ettinger_probing_2016,belinkov_what_2017,hupkes2018visualisation,conneau_what_2018}, which attempts to unearth underlying linguistic properties by fitting relatively underparameterised linear models 
over representations generated by an encoder.
These methods have also faced recent critique, for example, 
concerning the lack of transparency in the classifers' ability to \emph{extract} meaningful information, as opposed to \emph{learning} it. 
Alternative paradigms for interpretability have therefore been proposed, such as correlation-based methods \citep{raghu2017svcca,saphra2018understanding,kornblith_similarity_2019
,chrupala-alishahi-2019-correlating}. However, this critique does not invalidate diagnostic classification: indeed, more recent work \citep{hewitt-liang-2019-designing} describes methods to show the empirical validity of certain probes, via control tasks. 

Among probing studies specifically pertinent to our paper, 
\citet{blevins_deep_2018} demonstrate that deep RNNs are capable of encoding syntax given a variety of pre-training tasks, including language modeling. \citet{peters-etal-2018-dissecting} 
demonstrate that, regardless of 
encoder (recurrent, convolutional, or self-attentive), biLM-based pre-training results in similar high-quality representations that implicitly encode a variety of linguistic phenomena, layer by layer. 
Similarly, \citet{tenney2019bert} employ the `edge probing' approach of \citet{tenney_what_2019} 
to demonstrate that BERT implicitly learns the `classical NLP pipeline', 
with lower-level linguistic tasks 
encoded in lower layers 
and more complex phenomena 
in 
higher layers, 
and dependency syntax in layer 5--6.
Finally, \citet{hewitt2019structural} describe a syntactic probe for extracting aspects of dependency syntax from pre-trained representations
, which we describe 
in Section~\ref{sec:probe}.

\section{Aspects of Syntax}

Syntax studies how natural language encodes meaning using 
expressive devices such as word order, case marking and agreement. 
Some approaches 
emphasize the formal side and primarily try to account for the distribution of linguistic forms. 
Other frameworks 
focus on the functional side 
to capture the interface to semantics. 
And some theories use multiple representations to account for both perspectives, such as c-structure and f-structure in LFG \citep{kaplan82,bresnan00} or surface-syntactic and deep syntactic representations in Meaning-Text Theory \citep{melcuk88}.

When asking whether neural language models learn 
syntax, 
it is therefore relevant to ask which aspects of syntax we are concerned with. This is especially important 
if we probe the models by trying to extract syntactic representations, since these representations may be based on different theoretical perspectives. 
As a first step in this direction, we explore two different dependency-based syntactic representations, for which annotations are available in multiple languages. The first is Universal Dependencies (UD) \citep{nivre-etal-2016-universal}, a framework for cross-linguistically consistent morpho-syntactic annotation, which prioritizes direct grammatical relations between content words. These relations tend to be more parallel across languages that use different surface features to encode the relations. The second is Surface-Syntactic Universal Dependencies (SUD) \citep{gerdes18udw}, a recently proposed alternative to UD, which gives more prominence to 
function words in order to capture variations in surface structure across languages. 

Figure~\ref{fig:en-fi} contrasts the two frameworks by showing how they annotate an English sentence. While the two annotations agree on most syntactic relations (in black), including the analysis of core grammatical relations like subject (nsubj\footnote{UD uses the \emph{nsubj} relation, for \emph{nominal} subject, while SUD uses a more general \emph{subj} relation.}) and object (obj), they differ in the analysis of auxiliaries and prepositional phrases. The UD annotation (in blue) treats the main verb \emph{chased} as the root of the clause, while the SUD annotation (in red) assigns this role to the auxiliary \emph{has}. The UD annotation has a direct oblique relation between \emph{chased} and \emph{room}, treating the preposition \emph{from} as a case marker, while the SUD annotation has an oblique relation between \emph{chased} and \emph{from}, analyzing \emph{room} as the object of \emph{from}. The purpose of the UD style of annotation is to increase the probability of the root and oblique relations being parallel in other languages that use morphology (or nothing at all) to encode the information expressed by auxiliaries and adpositions. 
SUD is instead designed to bring out differences in surface structure in such cases.  

The 
different treatment of function words affects 
not only adpositions (prepositions and postpositions) and auxiliaries (including copulas), but also subordinating conjunctions and infinitive markers. 
Because of these systematic differences, dependency trees in UD tend to have 
longer average dependency length and smaller height\footnote{The \emph{height} of a tree is the length of the longest path from the root to a leaf (sometimes referred to as \emph{depth}).} 
than in SUD. 

\section{Probing Model}
\label{sec:probe}

To conduct our experiments, we 
make use of the structural probe proposed by
\citet{hewitt2019structural}, which 
is made up of two complementary components --- distance and depth. The former is an intuitive proxy for the notion of two words being connected by a dependency: any two words $w_{i}, w_{j}$ in a tree $T$ 
are neighbors 
if their respective distance in the tree amounts to $d_{T}(w_{i},w_{j})=1$. This metric can theoretically be applied to the vector space of any pretrained neural language model sentence encoding, which ouputs a set of vectors $S =
\mathbf{h}_{1}, ..., \mathbf{h}_{n}$ for a sentence
. In practice, however, the distance 
between any two vectors $\{\mathbf{h}_{i}, \mathbf{h}_{j}\} \in S$ will not be directly comparable to their 
distance in a corresponding syntactic tree $T$, 
because the model does not encode syntax in isolation. To resolve this, \citet{hewitt2019structural} propose to learn a linear transformation matrix $B$, such that $d_{B}(\mathbf{h}_{i}, \mathbf{h}_{j})$
extracts the 
distance between any two words $w_{i}, w_{j}$ in a 
parse tree. For an annotated corpus of $L$ sentences, the distance probe can be learned via gradient descent as follows:

\[
\min_{B}\sum_{l=1}^{L}
\frac{1}{|n^{l}|^{2}}\sum_{i,j}|d_{T^{l}}(w_{i}^{l}, w_{j}^{l}) -
d_{B}(\mathbf{h}_{i}^{l}, \mathbf{h}_{j}^{l})^{2}|
\]
\noindent
where $|n^{l}|$ is the length of sentence $l$, 
normalized by 
the number $|n^{l}|^{2}$ of word pairs, and $d_{T^{l}}(w_{i}^{l},
w_{j}^{l})$ is the 
distance of words $w_{i}^l$ and $w_{j}^l$ in the gold tree.

While the 
distance probe can 
predict which words enter into dependencies with one another, it is insufficient for predicting which 
word is the head. To resolve this, \citet{hewitt2019structural} employ a separate probe for tree depth,\footnote{The \emph{depth} of a node is the length of the path from the root.} where they make a similar assumption as they do for distance: a given (square) vector L2 norm
$||\mathbf{h}_{i}^{2}||$ is analogous to $w_{i}$'s depth in a tree $T$. A linear transformation matrix $B$ can therefore be learned in a similar way:
\[
\min_{B}\sum_{l=1}^{L}\frac{1}{n_{l}}\sum_{i}^{n}(||w_{i}^{l}|| -
||B\mathbf{h}_{i}^{l}||^2)
\]
where $||w_{i}^{l}||$ is the depth of a $w_{i}^{l}$ 
in the gold tree.

To be able to score probed trees (against UD and SUD gold trees) using the standard metric of unlabeled attachment score (UAS), we need to derive a rooted directed dependency tree from the information provided by the distance and depth probes. 
Algorithm \ref{alg:cle} outlines a simple method to retrieve a well-formed tree with the help of the Chu-Liu-Edmonds maximum spanning tree algorithm \citep{chu65,mcdonald05emnlp}. 
Essentially, in a sentence $S = w_{1} \dots w_{n}$, for every pair of nodes $(w_i, w_j)$ with an estimated distance of $d$ between them, if $w_i$ has smaller depth than $w_j$, we set the weight of the arc $(w_i, w_j)$ to $-d$; otherwise, we set the weight to $-\infty$. This is effectively a mapping from distances to scores, with larger distances resulting in lower arc scores from the parent to the child, and infinitely low scores from the child to the parent. We also add a pseudo-root $w_0$ (essential for decoding), which has a single arc pointing to the shallowest node (weighted 0). We use the AllenNLP~\citep{gardner2018allennlp} implementation of the Chu-Liu/Edmonds' algorithm. 


\begin{algorithm}[t]
\caption{Invoke CLE for sentence $S = w_{1,n}$\\ given distance matrix $E$ and depth vector $D$}
\begin{algorithmic}
\Procedure{InvokeCLE}{$E, D$}
\State{$N \gets |S| + 1$} 
\State{$M \gets \Call{Init}{shape\!=\!(N, N), value\!=\!-\infty}$}
\For{$(w_i, w_j) \in E$} 
    \If{$D(w_i) < D(w_j)$}
        \State{$M(w_{i}, w_{j}) \gets -E(w_i, w_j)$}
    \EndIf
\EndFor
\State{$root \gets \mathbf{argmin}_i D(w_i)$}
\State{$M(0, w_{root}) \gets 0$}
\State{\textbf{return} \Call{CLE}{$M$}}
\EndProcedure
\end{algorithmic}
\label{alg:cle}
\end{algorithm}

\begin{table*}[t]
\label{table:stats}
\small
\centering
\begin{tabular}{@{}lllrrrrrrrrr@{}}
\toprule
\textbf{Language} &\textbf{Code} &\textbf{Treebank}                   & \textbf{\#\ Sents} & \textbf{\%ADP} & \textbf{\%AUX} & \multicolumn{2}{c}{\textbf{\%ContRel}} & \multicolumn{2}{c}{\textbf{Dep Len}} & \multicolumn{2}{c}{\textbf{Height}} \\ \midrule
\multicolumn{3}{l}{} &          &          &          & UD               & SUD             & UD              & SUD             & UD              & SUD            \\ \midrule
Arabic & arb & PADT & 6075     & 15     & 1     & 37             & 24            & 4.17            & 3.92            & 7.20            & 9.82           \\
Chinese & cmn & GSD & 3997     & 5     & 3     & 37             & 30            & 3.72            & 3.74            & 4.30            & 6.56           \\
English & eng & EWT & 12543    & 8     & 6     & 20             & 12            & 3.13            & 2.94            & 3.48            & 5.11           \\
Basque & eus & BDT & 5396     & 2     & 13     & 34             & 25            & 2.99            & 2.90            & 3.49            & 4.18           \\
Finnish & fin & TDT & 12217    & 2     & 7     & 35             & 30            & 2.98            & 2.91            & 3.42            & 4.22           \\
Hebrew & heb & HTB & 5241     & 14     & 2     & 28             & 14            & 3.76            & 3.53            & 5.07            & 7.30           \\
Hindi & hin & HDTB & 13304    & 22     & 9     & 26             & 10            & 3.44            & 3.05            & 4.25            & 7.41           \\
Italian & ita & ISDT & 13121    & 14     & 5     & 21             & 8            & 3.30            & 3.12            & 4.21            & 6.28           \\
Japanese & jap & GSD & 7125     & 25     & 14     & 31             & 10            & 2.49            & 2.08            & 4.40            & 8.18           \\
Korean & kor & GSD & 4400     & 2     & 0     & 58             & 57            & 2.20            & 2.17            & 3.86            & 4.07           \\
Russian & rus & SynTagRus & 48814    & 10     & 1     & 31             & 22            & 3.28            & 3.13            & 4.21            & 5.24           \\
Swedish & swe & Talbanken & 4303     & 12     & 5     & 29             & 17            & 3.14            & 2.98            & 3.50            & 5.02           \\
Turkish & tur & IMST & 3664     & 3     & 2     & 33             & 30            & 2.21            & 2.12            & 3.01            & 3.37           \\ \midrule
Average & - & - & 10784.62 & 12 & 5 & 32	& 22 & 3.14 &	3.00	& 4.20	& 5.91 \\ \bottomrule
\end{tabular}
\caption{Treebank statistics: number of sentences (\# Sents) and percentage of adpositions (ADP) and auxiliaries (AUX). Comparison of UD and SUD: percentage of direct relations involving only nouns and/or verbs (ContRel); average dependency length (DepLen) and average tree height (Height).
Language codes are ISO 639-3.}
\label{tab:tb}
\end{table*}

\section{Experimental Design}

In order to evaluate the extent to which a given model's representational space fits either annotation framework, we fit the 
structural probe on the model, layer by layer, using 
UD and SUD treebanks 
for supervision, and compute UAS over each treebank's test set as a proxy for a given layer's goodness-of-fit. 

\paragraph{Language and Treebank Selection} 
We reuse the sample of \citet{kulmizev-etal-2019-deep}, which comprises 13 languages from different language families, with different morphological complexity, and with different scripts. 
We use treebanks from UD v2.4 \cite{ud24} and their conversions into SUD.\footnote{https://surfacesyntacticud.github.io/data/}
Table~\ref{tab:tb} shows background statistics for the treebanks, including the percentage of adpositions (ADP) and auxiliaries (AUX), two important function word categories that are treated differently by UD and SUD. A direct comparison of the UD and SUD representations shows that, as expected, UD has a higher percentage of relations directly connecting nouns and verbs (ContRel), higher average dependency length (DepLen) and lower average tree height (Height). However, the magnitude of the difference varies greatly across languages.\footnote{For Chinese, UD actually has slightly lower average dependency length than SUD.}

\paragraph{Models} We evaluate two pretrained language models: BERT \citep{devlin-etal-2019-bert} and ELMo \citep{peters-etal-2018-deep}. 
For BERT, we use the pretrained \texttt{multilingual-bert-cased} model provided by Google.\footnote{https://github.com/google-research/bert} The model is trained on the concatenation of WikiDumps for the top 104 languages with the largest Wikipedias and features a 12-layer Transformer with 768 hidden units and 12 self-attention heads. For ELMo, we make use of the pretrained monolingual models made available by \citet{che18}. These models are trained on 20 million words randomly sampled from the concatenation of WikiDump and CommonCrawl datasets for 44 different languages, including our 13 languages. Each model features a character-based word embedding layer, as well as 2 bi-LSTM layers, each of which is 1024-dimensions wide.

Though we fit the probe on all layers of each model separately, we also learn a weighted average over each full model: 

\[
\label{eq:1}
  \mathbf{model}_i = \sum_{j=0}^{L} s_{j} \mathbf{h}_{i,j}
\]
\noindent
where $s_j$ is a learned parameter, $\mathbf{h}_{i,j}$ is the encoding of word $i$ at layer $j$, and $L$ is the number of layers. We surmise that, in addition to visualizing the probes' fit across layers, this approach will give us a more general notion of how well either model aligns with the respective frameworks. We refer to this representation as the 13th BERT layer and the 3rd ELMo layer. When determining the dimensionality of the transformation matrix (i.e. probe rank), we defer to each respective encoder's hidden layer sizes. However, 
preliminary experiments indicated that probing accuracy was stable across ranks of decreasing sizes. 

It is important to note that by \textit{probe} we henceforth refer to the algorithm that combines both distance and depth probes to return a valid tree. One could argue that, per recent insights in the interpretability literature (e.g. \cite{hewitt-liang-2019-designing}), this model is too expressive in that it combines supervision from two different sources. We do not consider this a problem, as the two probes are trained separately and offer views into two different abstract properties of the dependency tree. As such, we do not optimize for UAS directly. 

\section{Results and Discussion}

\begin{figure*}[t!]
    \centering
    \includegraphics[scale=0.12]{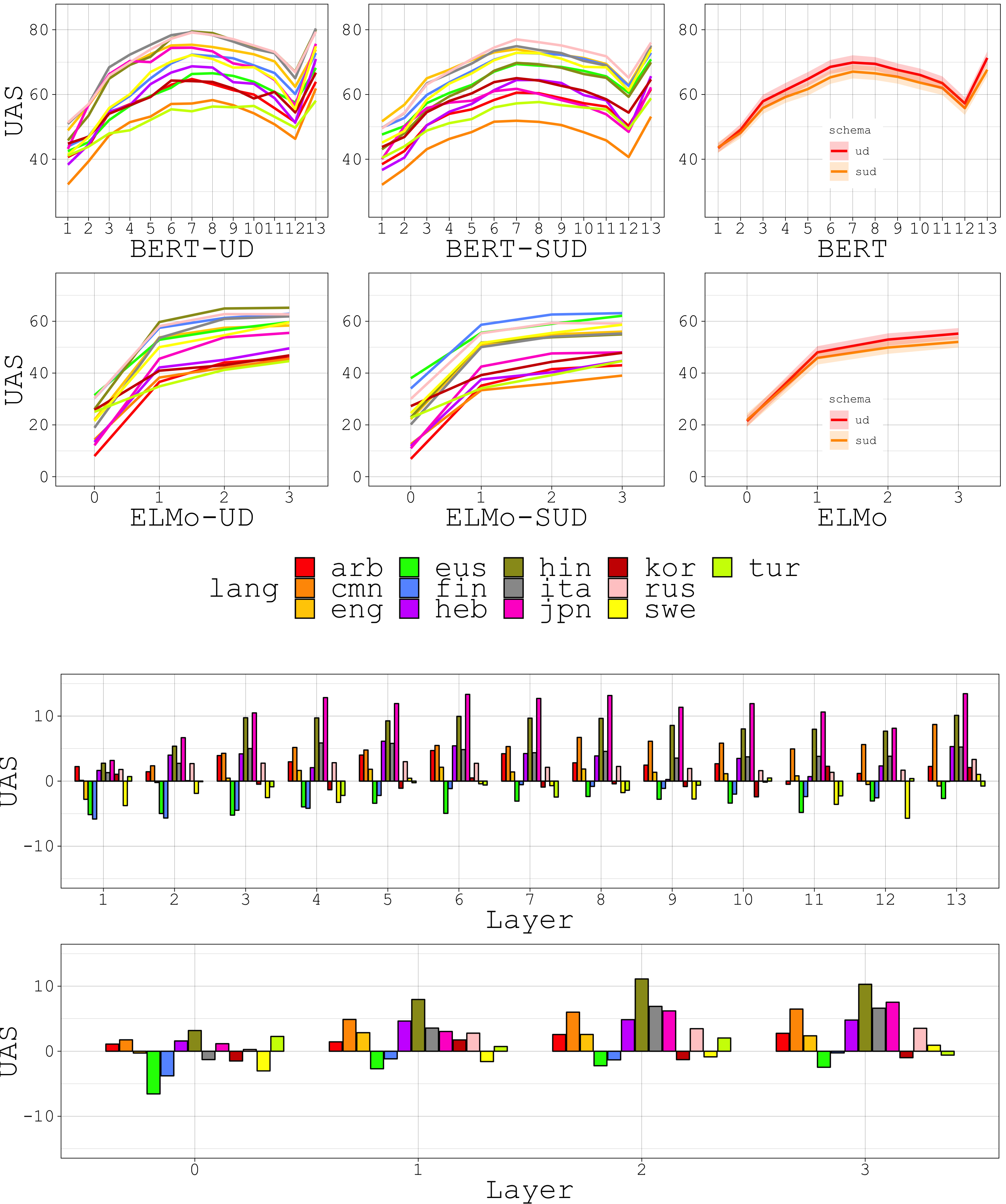}
    \caption{Probe results per model, layer, and language. First two rows depict UAS per layer and language for BERT and ELMo, with average performance and error over UD/SUD in 3rd column. Bottom two rows depict the difference in UAS across UD ($+$) and SUD ($-$) per model.}
    \label{fig:results}
\end{figure*}

Figure \ref{fig:results} displays the UAS after fitting the structural probes on BERT and ELMo, per language and layer. What is perhaps most noticeable is that, while BERT can achieve accuracies upwards of 79 UAS on some languages, ELMo fares consistently worse, maxing out at 65 for Hindi at layer 2. The most likely explanation for this is that the ELMo models are 
smaller than the multilingual BERT's 12-layer Transformer-based architecture, which was trained on orders of magnitude more data (albeit multilingually). 


In general, we find that the probing performance is stable across languages, where layers 7--8 fare the best for BERT and layer 2 for ELMo.\footnote{It is important to note that layer 0 for ELMo is the non-recurrent embedding layer which contains no contextual information.} This 
contrasts with prior observations \cite{tenney2019bert}, as the syntactic `center of gravity' is placed higher in each 
model's hierarchy. 
However, computing a weighted average over layers tends to produce the best overall performance for each model, 
indicating that the probe 
can benefit from 
information encoded across various layers. 

Once we compare the averaged results across syntactic representations, a preference for UD emerges, starting in layer 3 in BERT and layer 2 in ELMo. We observe the max difference in favor of UD in layer 7 for BERT, where the probe performs 3 UAS points better than SUD, 
and in the weighted average (layer 13), 
with 4 UAS points. 
The difference for the 13th BERT and 3rd ELMo layers is statistically significant at $p \leq 0.05$ (Wilcoxon signed ranks test).
A further look at 
differences across languages reveals that, while most languages tend to overwhelmingly prefer UD, there are some that do not: Basque, Turkish, and, to a lesser extent, Finnish. Furthermore, the preference towards SUD in these languages tends to be most pronounced in the first four and last two layers of BERT. However, in the layers where we tend to observe the higher UAS overall (7--8), this is minimized for Basque/Turkish and almost eliminated for Finnish. Indeed, we see the strongest preferences for UD in these layers overall, where Italian and Japanese are overwhelmingly pro-UD, to the order of 10+ UAS points. 

\subsection{Controlling for Treebank Size}

Overall, we note that some languages consistently achieve higher accuracy, like Russian with 71/69 UAS for UD/SUD for BERT, while others fare poorly, like Turkish (52/43) and Chinese (51/46). In the case of these languages, one can observe an obvious relation to the size of our reference treebanks, where Russian is by far the largest and Turkish and Chinese are the smallest. To test the extent to which training set size affects probing accuracy, we trained our probe on the same treebanks, truncated to the size of the smallest one --- Turkish, with 3664 sentences. Though we did observe a decline in accuracy in the largest treebanks (e.g. Russian, Finnish, and English) for some layers, the difference in aggregate was minimal. Furthermore, the magnitude of the difference in UD and SUD probing accuracy was almost identical to that of the probes trained on full treebanks, speaking to the validity of our findings. We refer the reader to Appendix \ref{sec:appendix_a} for these results. 

\subsection{Connection to Supervised Parsing}

Given that our findings seem to generally favor UD, another question we might ask is: are SUD treebanks simply harder to parse? This may seem like a straight-forward hypothesis, given SUD's tendency to produce higher trees in aggregate, which may affect parsing accuracy --- even in the fully supervised case. To test this, we trained UD and SUD parsers using the UDify model \citep{kondratyuk-straka-2019-75}, which employs a biaffine attention decoder \citep{dozat2016deep} after fine-tuning BERT representations (similar to our 13th layer). The results showed a slightly higher average UAS for UD (89.9 vs.\ 89.6) and a slightly higher LAS for SUD (86.8 vs.\ 86.5). Neither difference is statistically significant (Wilcoxon signed ranks test), which seems to rule out an alternative explanation in terms of learnability.  
We include the full range of results in Appendix \ref{sec:appendix_b}.

\begin{figure*}[t!]
    \centering
    \includegraphics[scale=0.12]{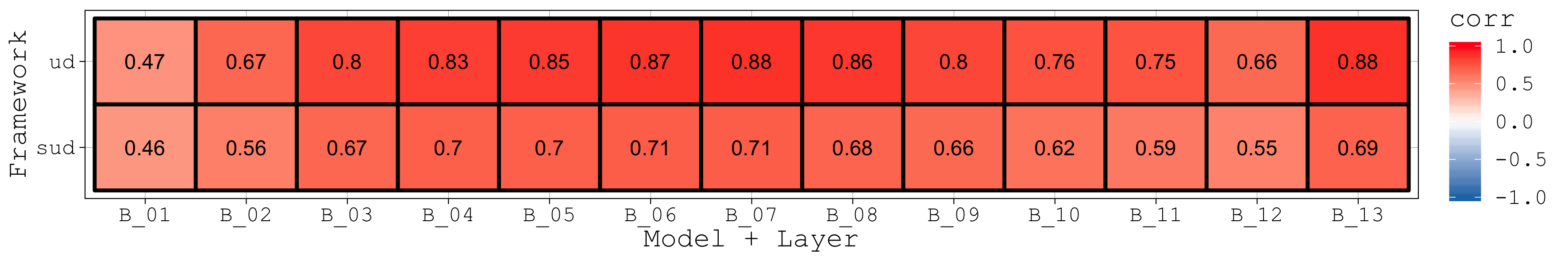}
    \caption{Pearson correlation between UD/SUD probing accuracy and supervised UAS, per layer.}
    \label{fig:corrs}
\end{figure*}

In addition to this, we tested how well each framework's probing accuracy related to supervised UAS across languages. We computed this measure by taking the Pearson correlation of each BERT probe's layer accuracy (per-language) with its respective framework accuracy. All correlations proved to be significant at $p \leq 0.05$, with the exception of UD and SUD at layer 1. Figure \ref{fig:corrs} displays these results. Here, we observe that probing accuracies correlate more strongly with supervised UAS for UD than for SUD. We can interpret this to mean that the rate at which trees are decoded by the UD probe is more indicative of how well they can be parsed given a full view of their structure, rather than vice-versa. Although correlation is an indirect measure here, we can still accept it to be in support of our general findings. 

\subsection{Parts of Speech}

\begin{figure}[t]
    \centering
    \includegraphics[scale=0.09]{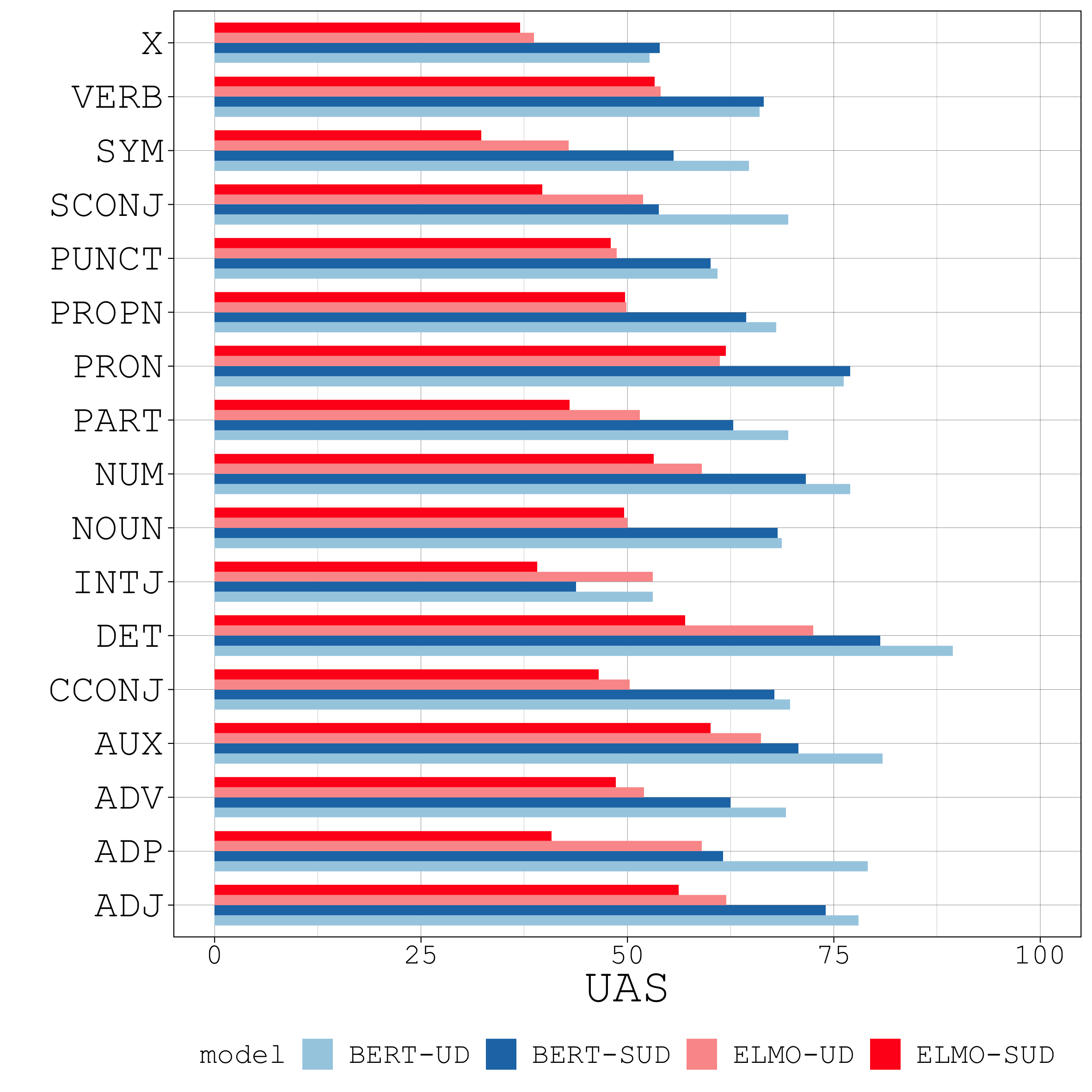}
    \caption{UAS accuracy for the average models (BERT 13, ELMo 3) on incoming dependencies of different part-of-speech categories.}
    \label{fig:pos_tag}
\end{figure}

In order to gain 
a better understanding of these probing patterns, we move on to an error analysis over the dev sets of each treebank, as fit by the 
averaged models. 
Figure \ref{fig:pos_tag} shows probe accuracy for different models (BERT/ELMo) and syntactic representations (UD/SUD) when attaching words of specific part-of-speech categories to their heads. The general pattern is that we observe higher accuracy for UD for both models on all categories, the only exceptions being a slightly higher accuracy for both models on PRON and for ELMo on VERB and X.\footnote{The X category is unspecified and extremely rare.} However, the differences are generally greater for function words, in particular ADP, AUX, SCONJ, PART and DET. In some respects, this is completely expected given the different treatment of these words in UD and SUD, and we can use the case of adpositions (ADP) to illustrate this. In UD, the preposition \emph{from} in a phrase like \emph{from the room} is simply attached to the noun \emph{room}, which is in general a short relation that is easy to identify. In SUD, the relation between the preposition and the noun is reversed, and the preposition now has to be attached to whatever the entire phrase modifies, which often means that difficult attachment ambiguities need to be resolved. However, exactly the same ambiguities need to be resolved for nominal words (NOUN, PRON, PROPN) in the UD representation, but there is no corresponding drop in accuracy for these classes in UD (except very marginally for PRON). Similar remarks can be made for other function word categories, in particular AUX, SCONJ and PART. It thus seems that the UD strategy of always connecting content words directly to other content words, instead of sometimes having these relations mediated by function words, results in higher accuracy overall when applying the probe to the representations learned by BERT and ELMo.

The behavior of different part-of-speech classes can also explain some of the differences observed across languages. In particular, as can be seen in Table~\ref{tab:tb}, most of the languages that show a clear preference for UD --- Chinese, Hebrew, Hindi, Italian and Japanese --- are all characterized by a high proportion of adpositions. Conversely, the three languages that exhibit the opposite trend --- Basque, Finnish and Turkish --- have a very low proportion of adpositions. The only language that does not fit this pattern is Chinese, which has a low percentage of adpositions but nevertheless shows a clear preference for UD. Finally, it is worth noting that Korean shows no clear preference for either representation despite having a very low proportion of adpositions (as well as other function words), but this is due to the more coarse-grained word segmentation of the Korean treebank, which partly incorporates function words into content word chunks.\footnote{This is reflected also in the exceptionally high proportion of direct content word relations; cf.\ Table~\ref{tab:tb}.} 

\begin{figure}[t]
    \centering
    \includegraphics[scale=0.09]{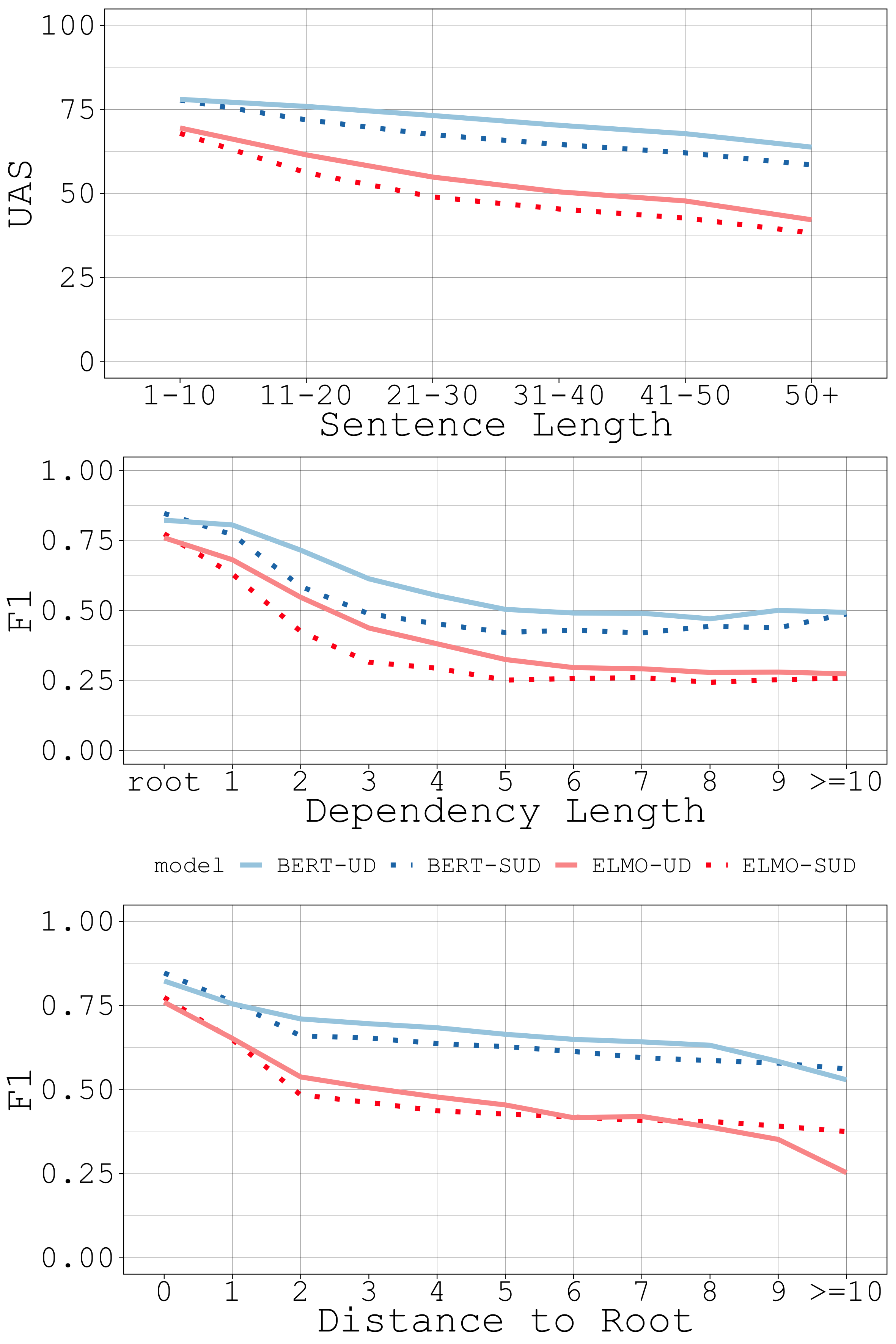}
    \caption{UAS across sentence length bins (top); F1 across varying dependency lengths (middle); F1 across varying distances to root (bottom)}
    \label{fig:graph}
\end{figure}

\subsection{Sentence and Tree Properties}

Figure \ref{fig:graph} depicts probing accuracy across different sentence lengths, dependency lengths, and distances to root. It is apparent that, despite the absolute differences between models, the relative differences between representations are strikingly consistent in favor of UD. For example, while the probe shows identical accuracy for the two representations for sentences of length 1--10, SUD decays more rapidly with increasing sentence length. Furthermore, while the SUD probe is slightly more accurate at detecting sentence roots and their immediate dependencies, we observe a consistent advantage for dependencies of length 2+, until dropping off for the longest length bin of 10+. Though Table \ref{tab:tb} indicates that UD dependencies are slightly longer than those of SUD, this factor does not appear to influence the probe, 
as there are no significant correlations between differences in average dependency length and differences in UAS.

\begin{figure*}[t]
    \centering
    \includegraphics[scale=0.095]{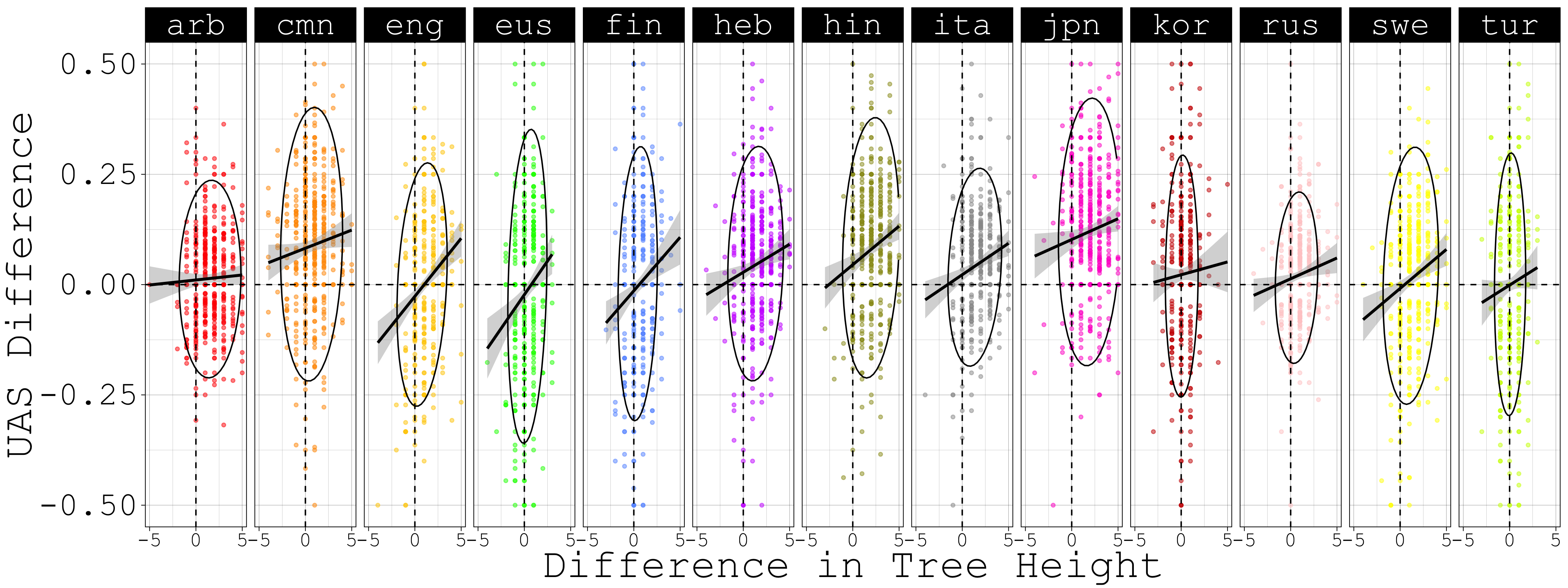}
    \caption{Differences in the BERT probe's UAS (UD $+$, SUD $-$) as a function of tree height per number of nodes (higher SUD tree $+$, higher UD tree $-$), with smoothed means and 95\% confidence ellipses as implemented in \texttt{ggplot2})}
    \label{fig:heights}
\end{figure*}

We observe a similar curve for varying distances to root, where the SUD probe performs slightly better than UD at the shortest distance, but decays faster for nodes higher in the tree. In general, UD trees have lower height than SUD (see Table~\ref{tab:tb}), which implies that tree height could be a major factor at play here. To verify this, we conducted a Pearson correlation test between the average increase in height from UD to SUD and the difference of the UD/SUD probe UAS per language. This test returned $\rho = 0.82, p < 0.001$, indicating that height is indeed crucial in accurately decoding trees across the two formalisms. In an attempt to visualize how this may play out across languages, we plotted the per-sentence difference in probing accuracy between UD/SUD as a function of the difference in height of the respective gold UD/SUD trees. Figure \ref{fig:heights} depicts these results for BERT, where the x-axis indicates how many nodes higher a SUD tree is with respect to its reference UD tree. 

It is apparent from Figure \ref{fig:heights} that the preference for UD can be largely explained via its lower tree height. If we first examine Korean, the segmentation of which results in the smallest difference in height overall, we observe a distribution that is roughly centered around zero on both axes. 
If we instead refer to the UD-preferring languages (Chinese, Hebrew, Hindi, Italian, and Japanese), we notice a strong skew of distributions towards the top right of the plot. This indicates (i) that the trees in these samples 
are higher for SUD and (ii) that the corresponding sentences are easier to decode in UD. By contrast, for the SUD-preferring languages (Basque, Finnish, and Turkish), we observe narrow distributions centered around 0 (similar to that of Korean), indicating minimal variation in tree height between UD and SUD. 
What these language have in common is an agglutinative morphology, which means that they rely more on morphological inflection to indicate relationships between content words, rather than separate function words. Sentences in these languages are therefore less susceptible to variations in tree height, by mere virtue of being shorter and possessing fewer relations that are likely be a better fit for UD, like those concerning adpositions. We speculate that it is this inherent property that explains the layerwise preference for SUD (though a general indifference in aggregate), allowing for some language-specific properties, like the crucial role of auxiliaries in Basque, to be easier to probe for in SUD. Conversely, with this in mind, it becomes easy to motivate the high preference for UD across some languages, given that they are not agglutinating and make heavy use of function words.
If we take the probe to be a proper decoding of a model's representational space, the encoding of syntactic structure according to an SUD-style analysis then becomes inherently more difficult, as the model is required to attend to 
hierarchy between words higher in the tree. 
Interestingly, however, this does not seem to correspond to an increased difficulty in the case of supervised parsing, as observed earlier.

\section{Conclusion and Future Work}

We have investigated the extent to which the 
syntactic structure captured by neural language models aligns with different styles of analysis, using UD treebanks and their SUD conversions as proxies. 
We have extended the structural probe of \citet{hewitt2019structural} to extract directed, rooted trees and fit it on pretrained BERT and ELMo representations for 13 languages. Ultimately, we observed a better overall fit for the UD-style formalism across models, layers, and languages, with some notable exceptions. For example, while the Chinese, Hebrew, Hindi, Italian, and Japanese models proved to be overwhelmingly better-fit for UD, Basque aligned more with SUD, and Finnish, Korean and Turkish did not exhibit a clear preference. Furthermore, an error analysis revealed that, when attaching words of various part-of-speech tags to their heads, UD fared better across the vast majority of categories, most notably adpositions and determiners. Related to this, we found a strong correlation between differences in average tree height and the tendency to prefer one framework over the other. This suggested a tradeoff between morphological complexity --- where differences in tree height between UD and SUD are minimal and probing accuracy similar --- and a high proportion of function words --- where SUD trees are significantly higher and probing accuracy favors UD. 

For future work, besides seeking a deeper understanding of the interplay of linguistic factors and tree shape, we want to explore probes
that combine the distance and depth assumptions into a single transformation, rather than learning separate probes and combining them post-hoc, 
as well as methods for alleviating treebank supervision altogether. 
Lastly, given recent criticisms of probing approaches in NLP, it will be vital to revisit the insights produced here within a non-probing framework, 
for example, using 
Representational Similarity Analysis (RSA) \citep{chrupala-alishahi-2019-correlating} over symbolic representations from treebanks and their encoded representations. 

\subsection*{Acknowledgements}

We want to thank Miryam De Lhoneux, Paola Merlo, Sara Stymne, and Dan Zeman and the ACL reviewers and area chairs for valuable feedback on preliminary versions of this paper. We acknowledge the computational resources provided by CSC in Helsinki and Sigma2 in Oslo through NeIC-NLPL (\texttt{www.nlpl.eu}).

\bibliography{acl2020}
\bibliographystyle{acl_natbib}
\clearpage
\appendix
\onecolumn
\section{Controlling for Treebank Size}
\label{sec:appendix_a}

\begin{figure}[ht!]
    \centering
    \includegraphics[scale=0.12]{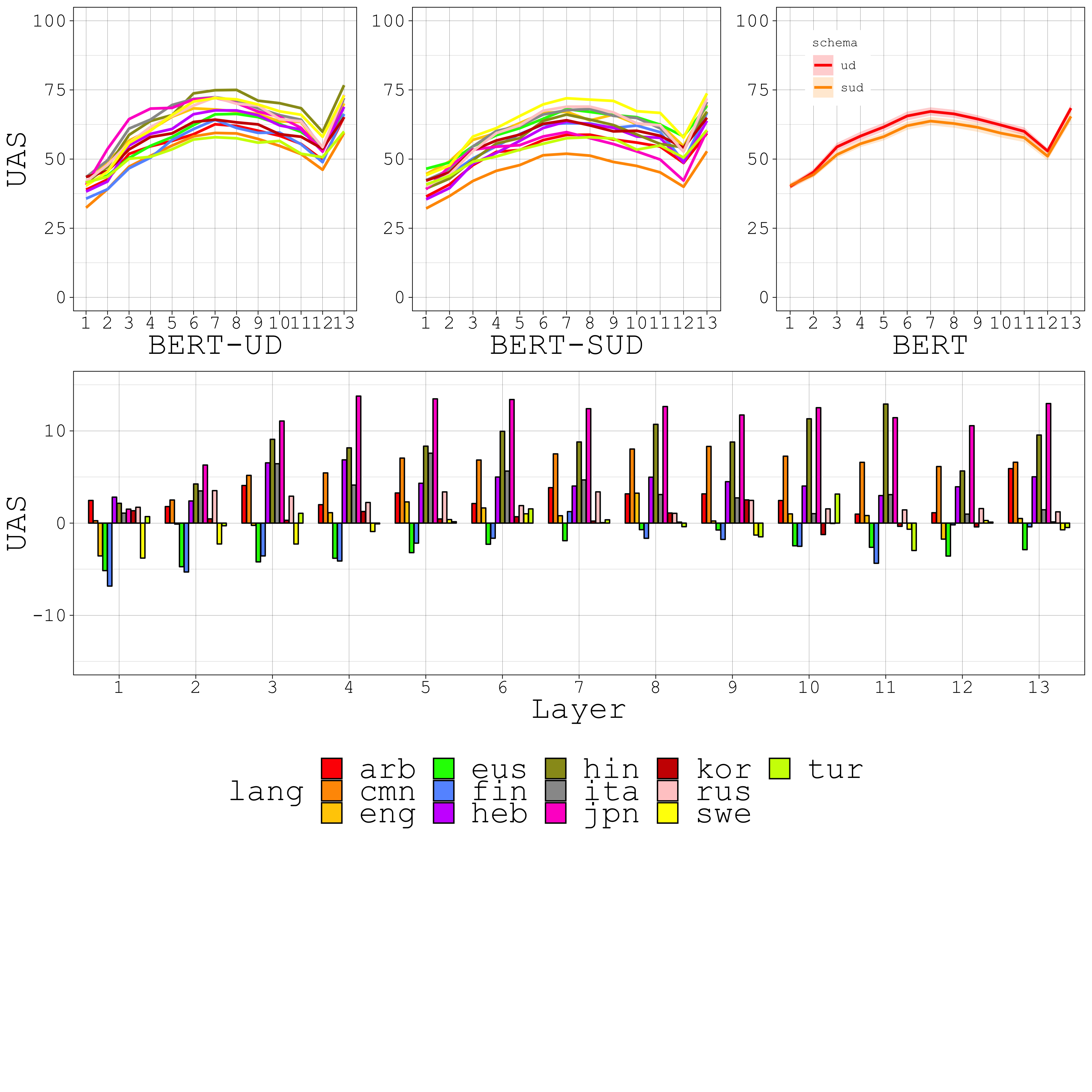}
    \caption{Probe results per framework, layer, and language, when trained on 3664 sentences. First row depicts UAS per layer and language for BERT, with average performance and error over UD/SUD in 3rd column. Bottom two row depicts the difference in UAS across UD ($+$) and SUD ($-$).}
    \label{fig:trunc1}
\end{figure}

\begin{figure*}[ht!]
    \centering
    \includegraphics[scale=0.12]{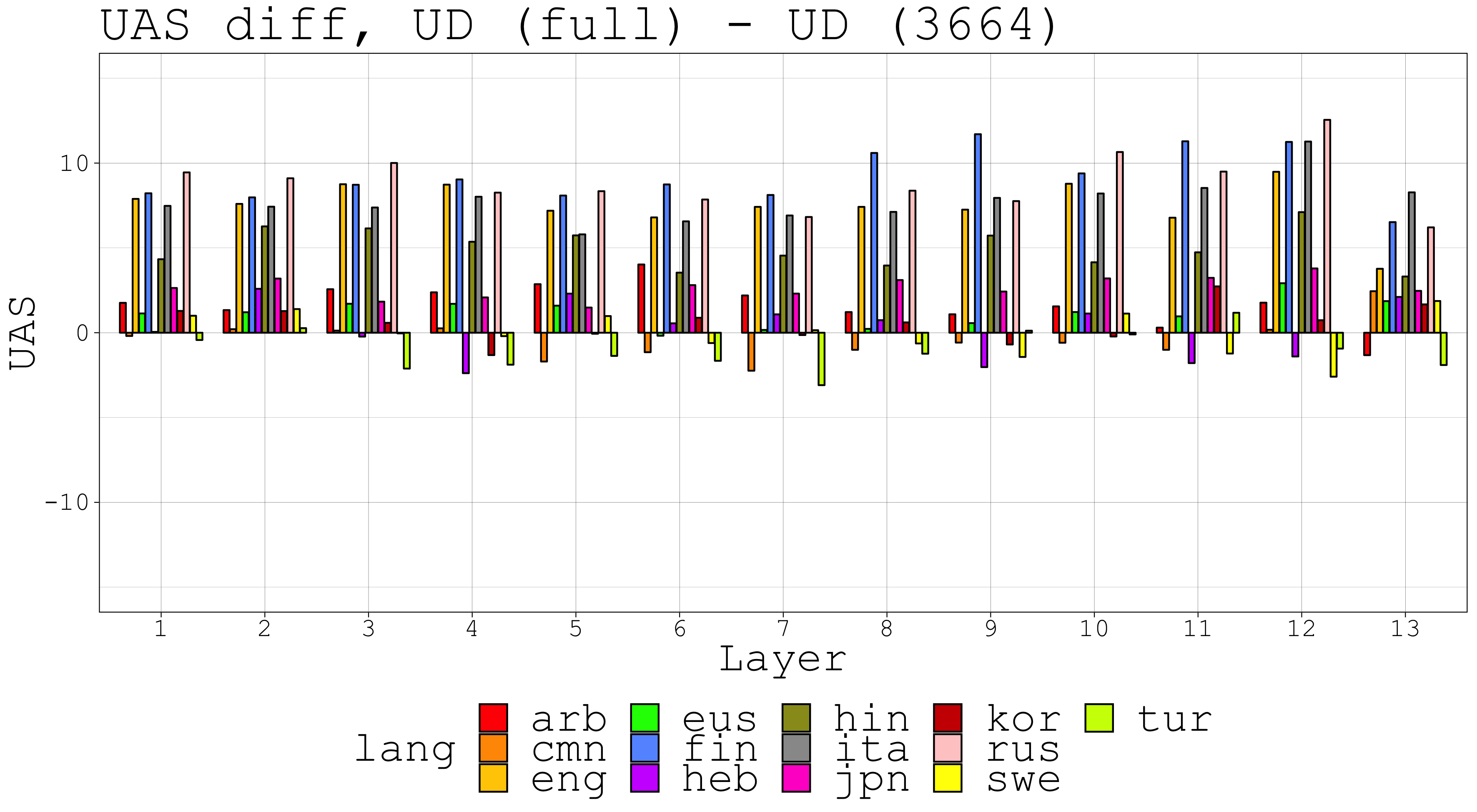}
    \caption{Difference in UAS across the UD probes trained on full data ($+$) and 3664 sentences ($-$).}
    \label{fig:trunc2}
\end{figure*}

\begin{figure*}[ht!]
    \centering
    \includegraphics[scale=0.12]{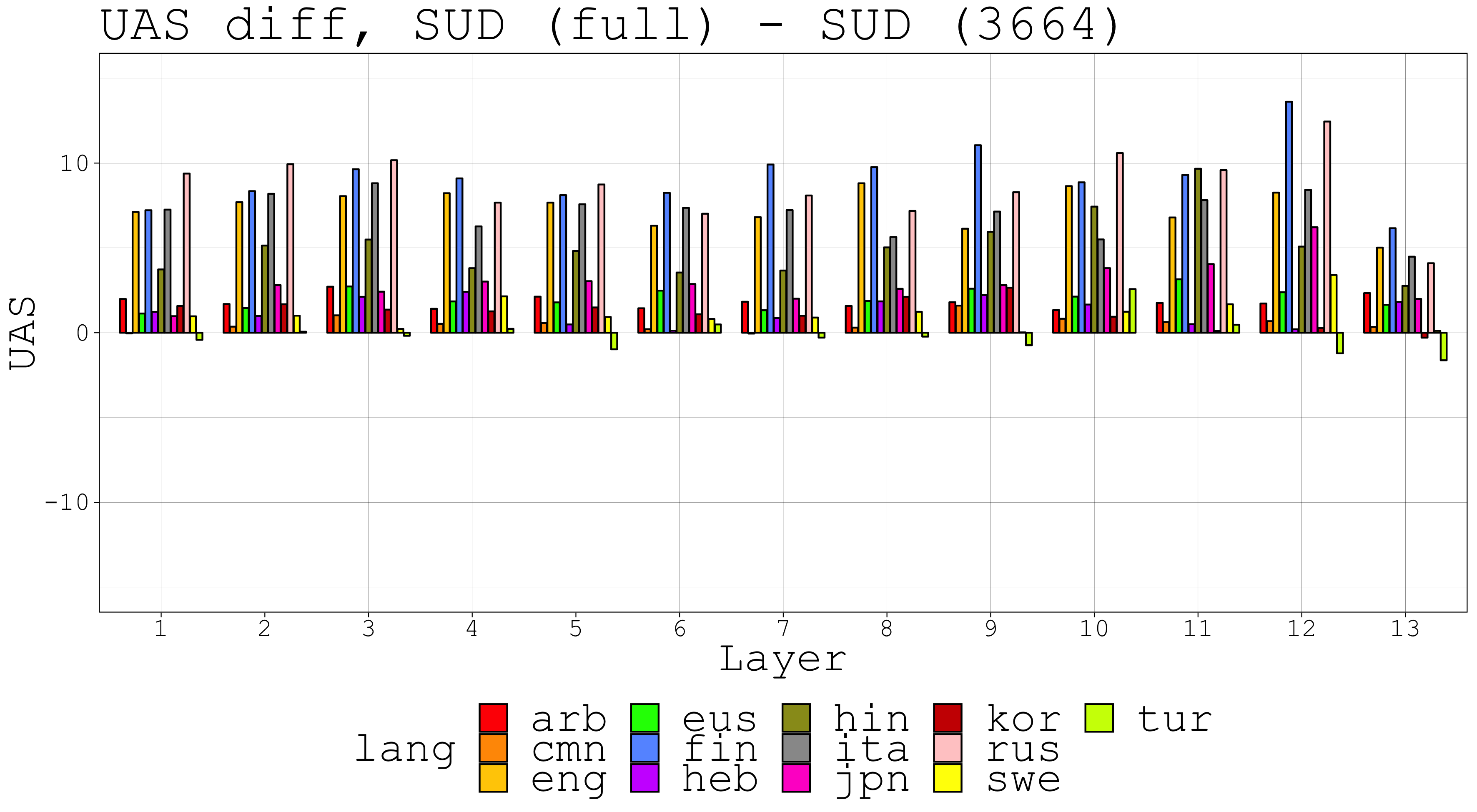}
    \caption{Difference in UAS across the SUD probes trained on full data ($+$) and 3664 sentences ($-$).}
    \label{fig:trunc3}
\end{figure*}
\clearpage
\section{Connection to Supervised Parsing}
\label{sec:appendix_b}

\begin{figure*}[ht!]
    \centering
    \includegraphics[scale=0.12]{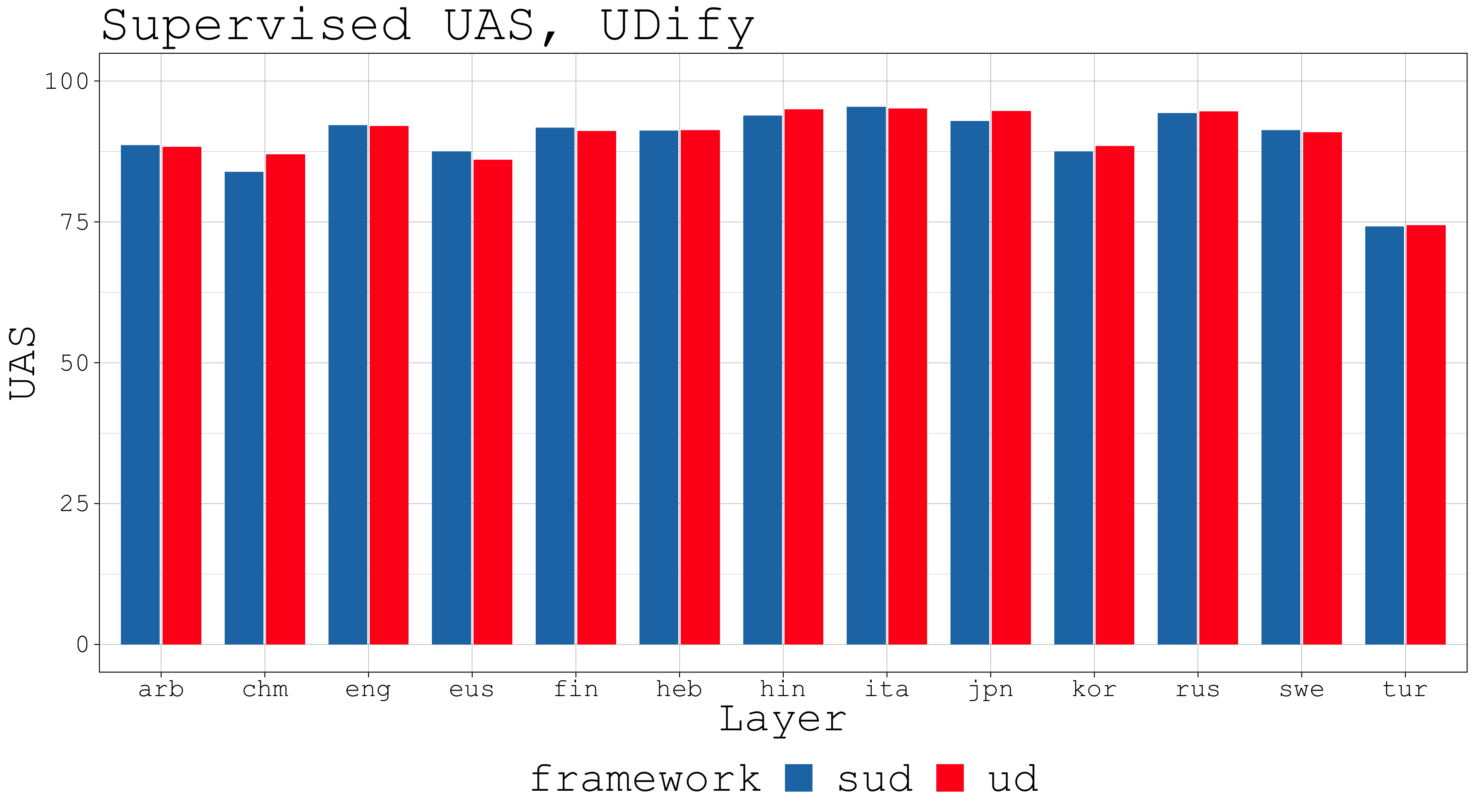}
    \caption{Supervised UDify UAS, UD and SUD, for all languages.}
    \label{fig:super1}
\end{figure*}

\begin{figure*}[ht!]
    \centering
    \includegraphics[scale=0.12]{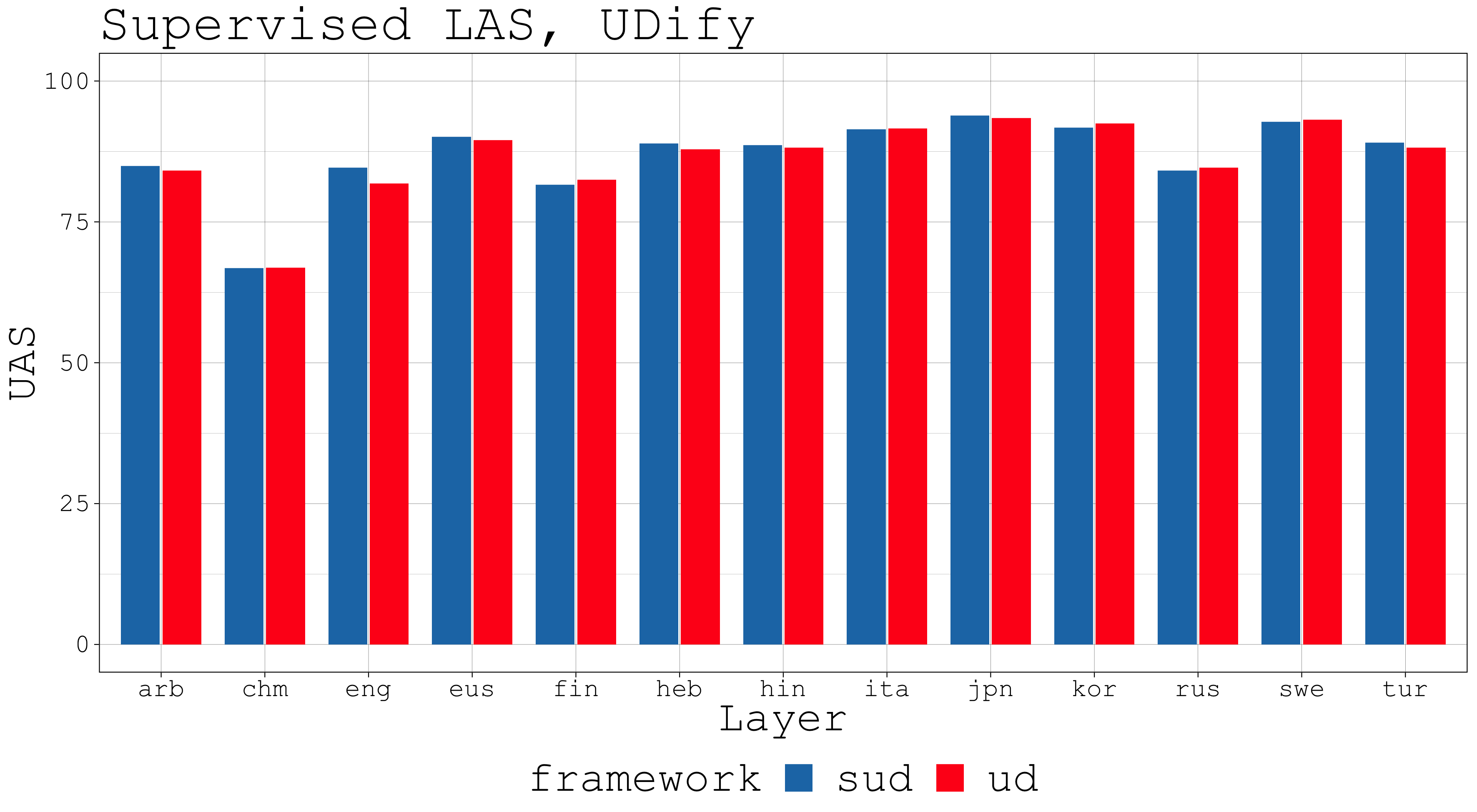}
    \caption{Supervised Udify LAS, UD and SUD, for all languages.}
    \label{fig:super2}
\end{figure*}

\end{document}